\documentclass[11pt]{article}

\usepackage{graphicx}
\graphicspath{{Figures/}}
\usepackage{subcaption}
\usepackage{tabulary}
\usepackage{multirow}
\usepackage[margin=1in]{geometry}
\usepackage{titlesec}
\usepackage[labelfont=bf,textfont=it,font=footnotesize]{caption}
\usepackage{parskip}
\usepackage[comma,sort&compress,numbers]{natbib}
\usepackage{amsmath}
\usepackage{mathpazo}
\usepackage{ulem}
\graphicspath{ {./figures/} }

\usepackage[pdfborder={0 0 0},colorlinks,allcolors=blue]{hyperref}

\titlespacing\section{0pt}{6pt plus 2pt minus 2pt}{6pt plus 2pt minus 2pt}
\titlespacing\subsection{0pt}{6pt plus 2pt minus 2pt}{6pt plus 2pt minus 2pt}
\titlespacing\subsubsection{0pt}{2pt plus 0pt minus 0pt}{2pt plus 0pt minus 0pt}
\titleformat{\section}{\large\bfseries\sffamily}{\thesection}{1em}{}
\titleformat{\subsection}{\normalsize\bfseries\sffamily}{\thesubsection}{1em}{}
\titleformat{\subsubsection}{\small\sffamily}{\thesubsubsection}{1em}{}

\newcommand{\coloredref}[2]{\hyperref[#2]{#1~\ref*{#2}}}
\newcommand{\comment}[1]{}
\newcommand{\ihat}{\hat \imath}

\begin{document}

\begin{center}
{\usefont{OT1}{phv}{b}{sc}\selectfont\Large{SpecXAI - Spectral interpretability of Deep Learning Models}}

{\usefont{OT1}{phv}{}{}\selectfont\normalsize
\vspace{0.1in}
{Stefan Druc$^1$, Peter Wooldridge$^1$, Adarsh Krishnamurthy$^2$, Soumik Sarkar$^2$,  Aditya Balu$^2$ \\
\vspace{0.1in}
$^1$Monolith AI LTD., London, England \\
$^2$Iowa State University, Ames, IA, USA 
\vspace{0.1in}
}}
\end{center}

\begin{abstract}
    Deep learning is becoming increasingly adopted in business and industry due to its ability to transform large quantities of data into high-performing models. These models, however, are generally regarded as black boxes, which, in spite of their performance, could prevent their use. In this context, the field of e\textbf{X}plainable \textbf{AI} attempts to develop techniques that temper the impenetrable nature of the models and promote a level of understanding of their behavior. Here we present our contribution to XAI methods in the form of a framework that we term SpecXAI, which is based on the spectral characterization of the entire network. We show how this framework can be used to not only understand the network but also manipulate it into a linear interpretable symbolic representation. 
\end{abstract}

\section{Introduction}
\label{sec:intro}
Deep learning has become a ubiquitous, versatile, and powerful technique that has a wide range of applications across many different fields such as image and speech recognition, natural language processing, and self-driving cars. The most popular application of deep learning is in the area of computer vision, where deep learning models are used for vision tasks such as image classification, object detection, and segmentation. 

While effective and powerful, one of the challenges that is plaguing deep learning models is explainability~\citep{xu2019explainable,dovsilovic2018explainable,adadi2018peeking,holzinger2022explainable}. Unlike traditional machine learning models, which can be understood through the use of simple mathematical equations, deep learning models are highly complex and difficult to interpret. This makes it difficult to understand how the model arrived at a particular decision, which can be a problem in areas such as healthcare~\citep{loh2022application}, or finance~\citep{vcernevivciene2022review,misheva2021explainable} where transparency is important. Explainability is also equally important for building trust, especially in several engineering applications that require Verification and Validation such as designing automotive vehicles and aircraft etc.~\citep{shukla2020opportunities,krishnamurthy2020explainable,srinivasan2021methodology,ogrezeanu2022privacy}.

Over the past decade, researchers have been working to develop methods to increase the explainability of deep learning models, such as through the use of visualization tools and interpretable representations of the model's internal workings~\citep{xu2019explainable,dovsilovic2018explainable,adadi2018peeking,holzinger2022explainable}. Despite these challenges, deep learning models have proven to be highly effective in a wide range of applications and they will continue to play a critical role in the field of artificial intelligence.

There are several explainability algorithms that have been proposed to help understand the decision-making process of deep learning models. A common approach is to try to linearize the model locally near the sample space to provide explanations.

One popular algorithm that uses this approach is LIME (Local Interpretable Model-Agnostic Explanations)~\cite{Ribeiro2016}. LIME approximates the complex deep learning model with a simpler, interpretable model that is only valid in the neighborhood of a specific sample. This allows us to understand the decision-making process of the deep learning model for that specific sample. LIME does this by perturbing the input data and measuring the change in the model's predictions. This information is then used to generate a linear model that approximates the complex model locally.

Another algorithm that uses this approach is SHAP (SHapley Additive exPlanations)~\cite{CASTRO20091726,trumbelj2010AnEE,Lundberg2017,giudici2021shapley}. SHAP is based on the concept of Shapley values from cooperative game theory, it provides an explanation for a specific sample by computing the contribution of each feature to the model's prediction. SHAP assigns a weight to each feature, which represents the importance of that feature in the final prediction. The weights are computed based on the feature's interactions with other features and the sample's proximity to the training data.

These algorithms are not the only ones that are available and other approaches like Anchors~\citep{alufaisan2021does}, Counterfactuals~\citep{keane2020good}, and Attention~\citep{liu2022rethinking,wiegreffe2019attention,zhang2022explainable} mechanisms can also be used to explain DL models. The main idea behind these algorithms is to linearize the model locally around the sample space to provide an explanation, which can be more interpretable and understandable than the complex deep learning model itself.

In contrast to these methods, several works have made use of the fact that neural networks can be represented as Point-wise Affine (PWA) Maps  around the input to explore the learned linear regions \cite{montfar2014number,Balestriero2021,lee2019robust,zhang2020empirical}. In other words, it has been shown that these complex models are in fact already locally linear and can be used directly rather than building linear surrogate models to explain the local behavior. \cite{10.5555/3045390.3045467,Lapuschkin_2019} create explanations of the model that can be attempted by analyzing the spectral decomposition of these linear representations across the dataset. In addition, spectral methods have been extensively applied to the weights of individual layers, most notably for compression and performance characterization. However, to our knowledge, until now there has been no attempt in the literature to analyze the spectral properties of the whole network via their PWA representation. 

In this paper, we view the network as an input-dependent operator and our explanations are generated by applying the SVD to decompose its action into more interpretable components. This operator view is also adopted in several works that attempt to learn orthogonal modes for fluid dynamics~\cite{EIVAZI2022117038}. These investigations, including our own, were in turn inspired by established dimensionality reduction techniques for dynamical systems such as the Proper Orthogonal Decomposition (POD)~\cite{Karhunen1946ZurSS} and the Dynamic Mode Decomposition (DMD)~\cite{rowley2009,Schmid2008}.

The main contributions of our work are:
\begin{enumerate}
    \item We make use of the Singular Value Decomposition (SVD) to understand the local representation of the network.
    \item We obtain linear symbolic expressions of the network output
    \item We use the singular vectors (SV) from multiple examples across the dataset to understand the global properties of the network.
\end{enumerate}






\section{Results}
\label{sec:res}


In this section, we present results by applying the above methodology to various models organized by dataset and task. 

\subsection{ILSVRC 2012 (ImageNet) Classification}

\begin{figure}[h!]
\centering
\begin{subfigure}{0.45\linewidth}
  \centering
  \includegraphics[width=\linewidth]{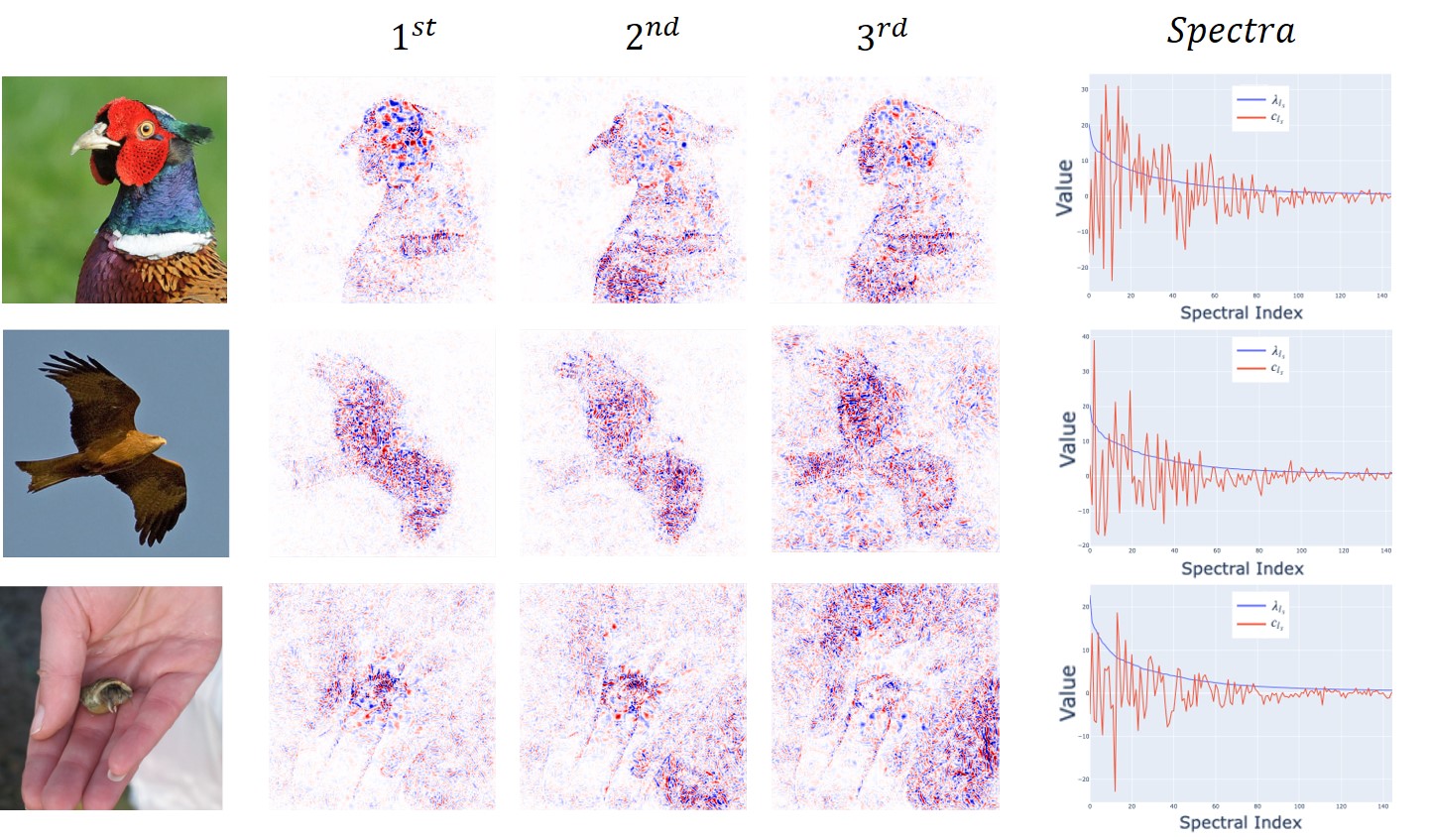}
  \caption{}
  \label{fig:c5_AN}
\end{subfigure}\hfil
\begin{subfigure}{0.45\linewidth}
  \centering
  \includegraphics[width=\linewidth]{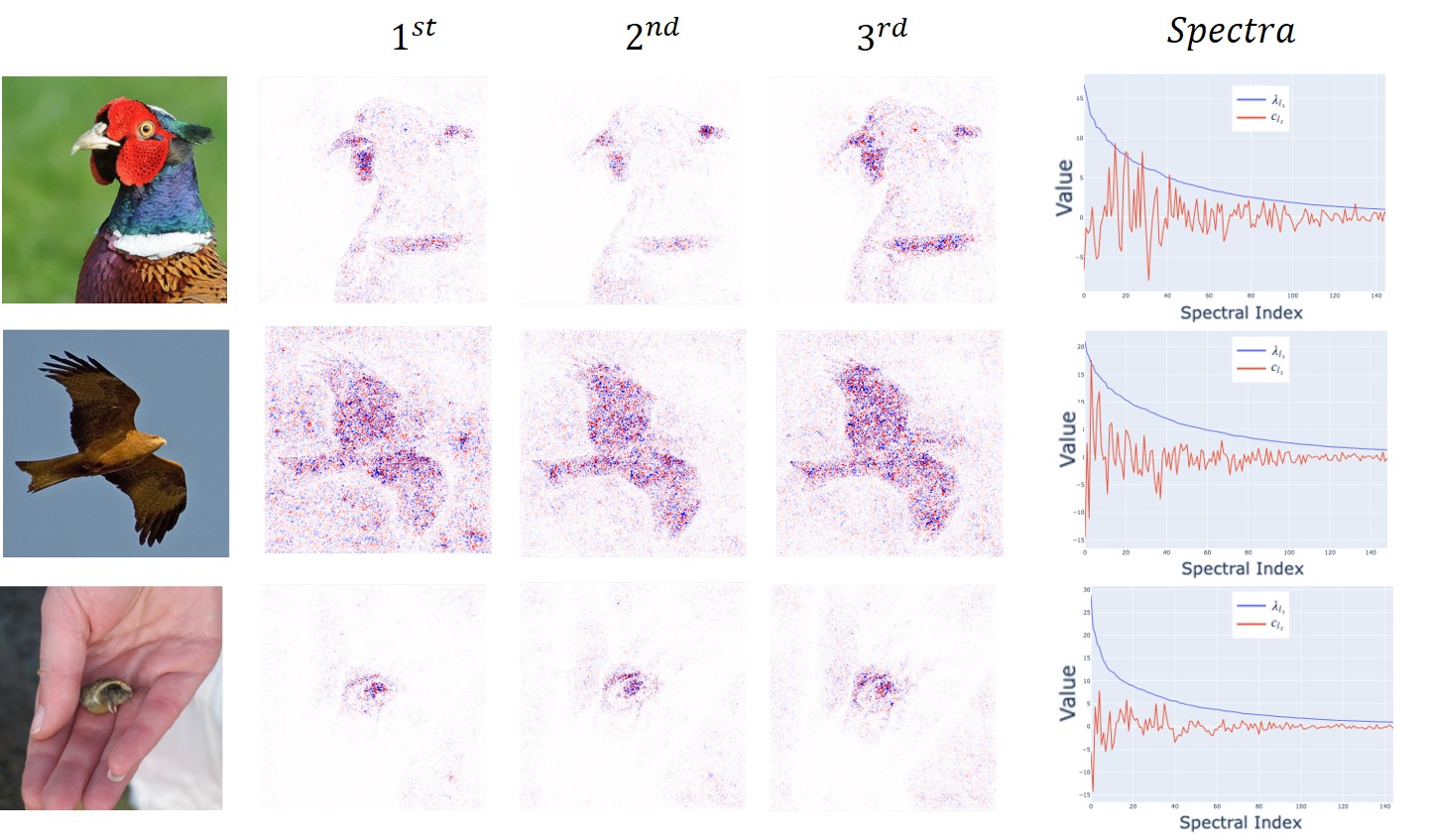}
  \caption{}
  \label{fig:c5_VGG}
\end{subfigure}
\caption{Comparison of top 3 SV's (in descending order) and spectra for (a) AlexNet and (b) VGG11 from 'classifier.5' layer}
\label{fig:c5_top_SVs}
\end{figure}

ImageNet~\cite{ILSVRC15} is a classification dataset consisting of roughly 1000 images for each of the 1000 possible classes (total of 1.2 million images). As one of the first large and well-curated image data sets, it has served as a standard benchmark to evaluate the state-of-the-art vision models. Here we examine two models trained on ImageNet. The first is the AlexNet~\citep{krizhevsky2017imagenet} CNN model which is credited to have started the deep learning boom in computer vision. The second is VGG11~\citep{simonyan2014very} with batch normalization. We use the pre-trained version of the models available through PyTorch and for both models we choose to explain the last layer before the output, i.e. $\text{'classifier.5'}$ layer. We present the feature-wise dot product for the top three SV's of a few examples for both models in Figure \ref{fig:c5_top_SVs}. We can note that for both models the SV's are mostly focused on the central object, however, the background becomes more prominent as the singular value decreases. In addition, VGG seems somewhat less focused on the background at this layer. 

\begin{figure}[ht!]
\centering
\begin{subfigure}{0.45\linewidth}
  \centering
  \includegraphics[width=\linewidth]{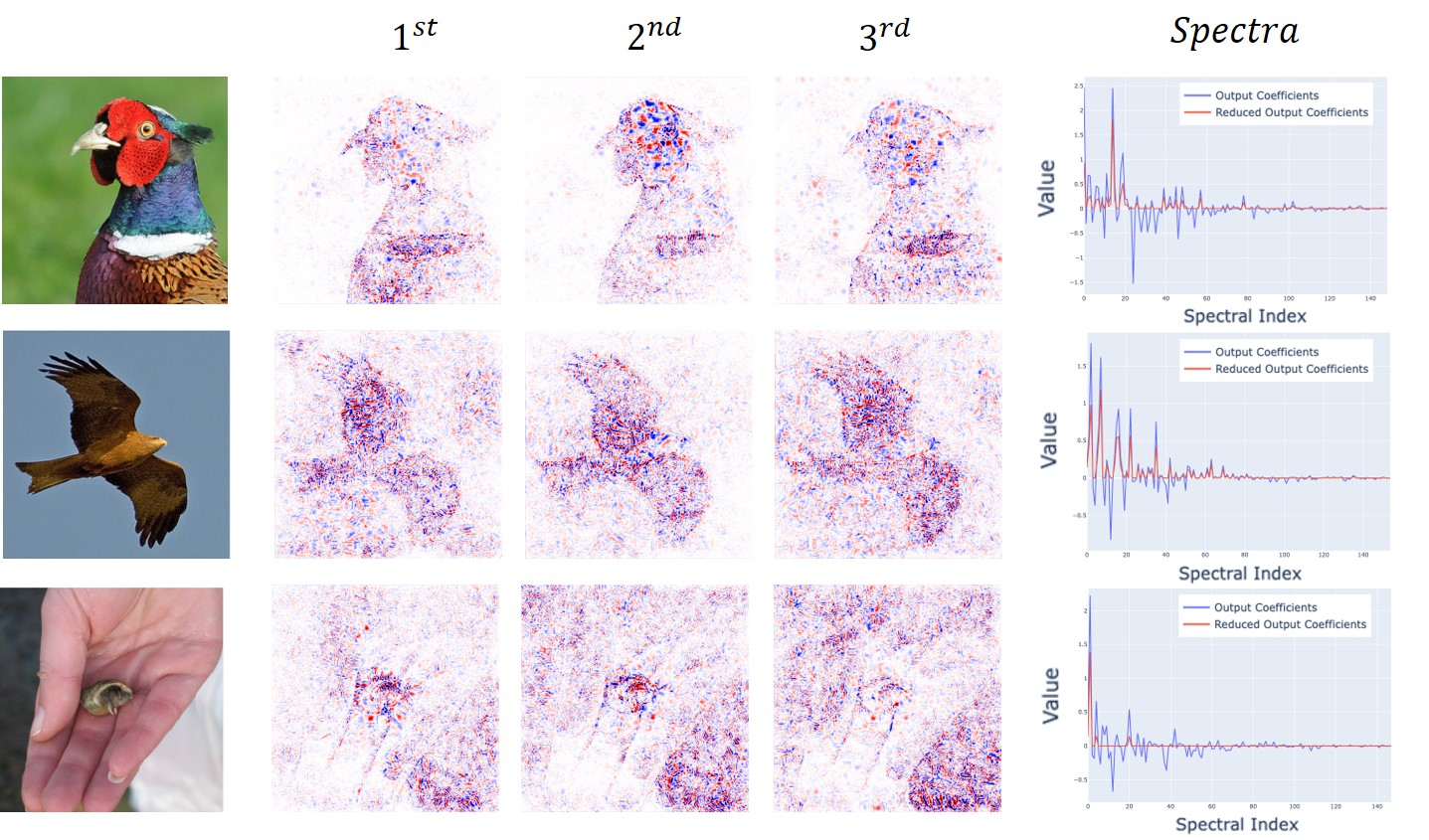}
  \caption{}
  \label{fig:o_c5_AN}
\end{subfigure}\hfil
\begin{subfigure}{0.45\linewidth}
  \centering
  \includegraphics[width=\linewidth]{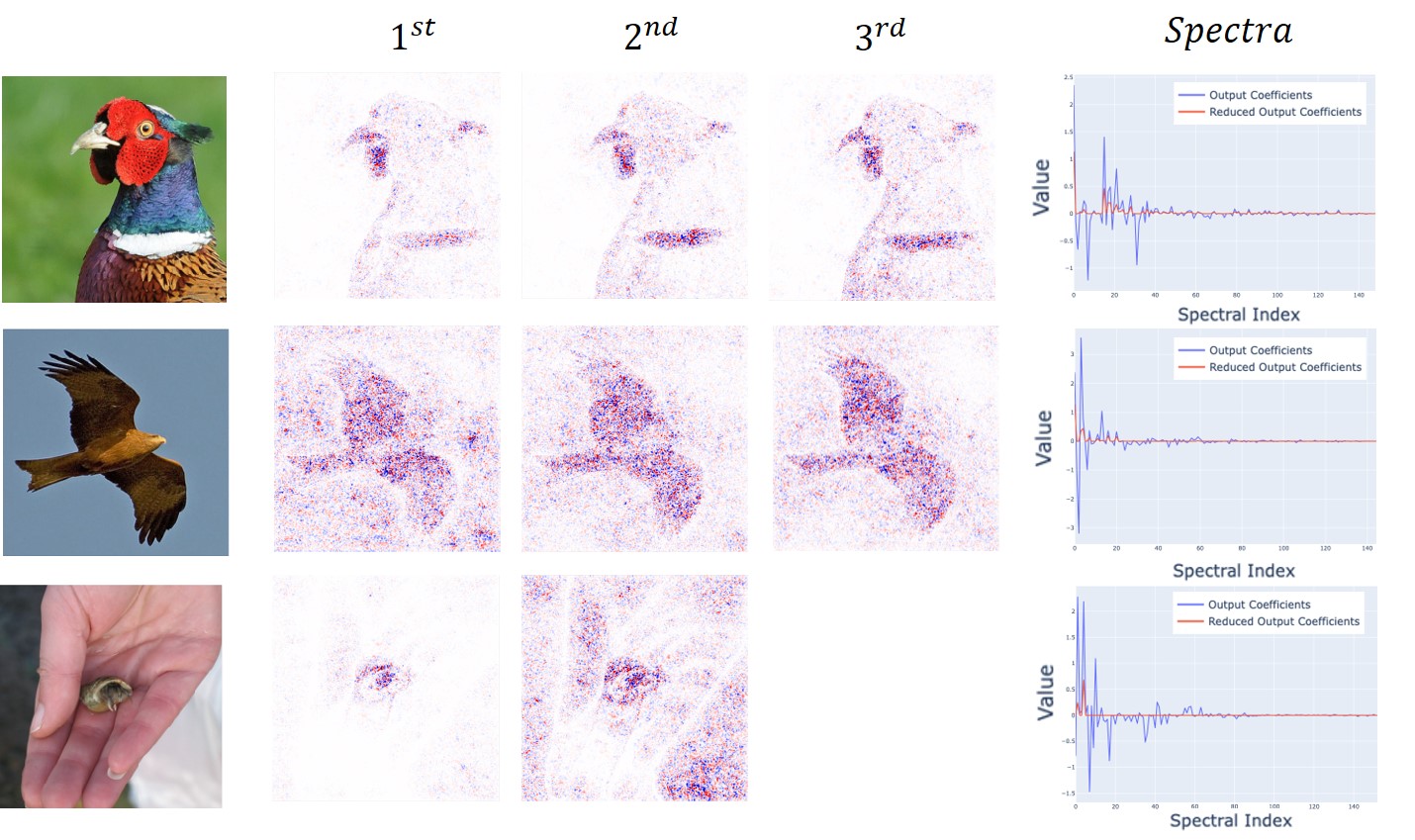}
  \caption{}
  \label{fig:o_c5_VGG}
\end{subfigure}
\caption{Comparison of top 3 SV's contributing to output (in descending order) for (a) AlexNet and (b) VGG11 from 'classifier.5' layer}
\label{fig:o_c5_top_SVs}
\end{figure}

It is important to note however that the order of singular vectors is not necessarily reflective of their contribution to the output. In Figure \ref{fig:o_c5_top_SVs}, we display the top contributing SV's to the output (the predicted class) after canceling coefficients to leave only positive contributions, i.e. $\hat{a}^j_{l_s}$ from (\ref{ahat}). Both models use noisier SV's from further down the spectrum, however, the reduced spectrum of AlexNet seems to be broader than VGG at this layer. Next, we proceed to explore the singular vectors generated from the last convolutional layer before global average pooling, specifically 'features.12' and 'features.28' for AlexNet and VGG respectively.   

The most noticeable change is that for both models the SV's seem to be more spatially localized on significant features, e.g. the wattle and the distinctive plumage of the fowl. In addition, the reduced spectra are generally sparser, e.g. the hermit crab only has a single SV contributing for AlexNet. The notable exception here is the eagle image for VGG11 which has a very broad reduced spectrum with no dominating SV's. All in all, this might suggest that the last convolutional layer is a better choice than the penultimate 'classifier.5' layer since it produces more intuitive and localized SV's, although not universally.

\begin{figure}[ht!]
\centering
\begin{subfigure}{0.45\linewidth}
  \centering
  \includegraphics[width=\linewidth]{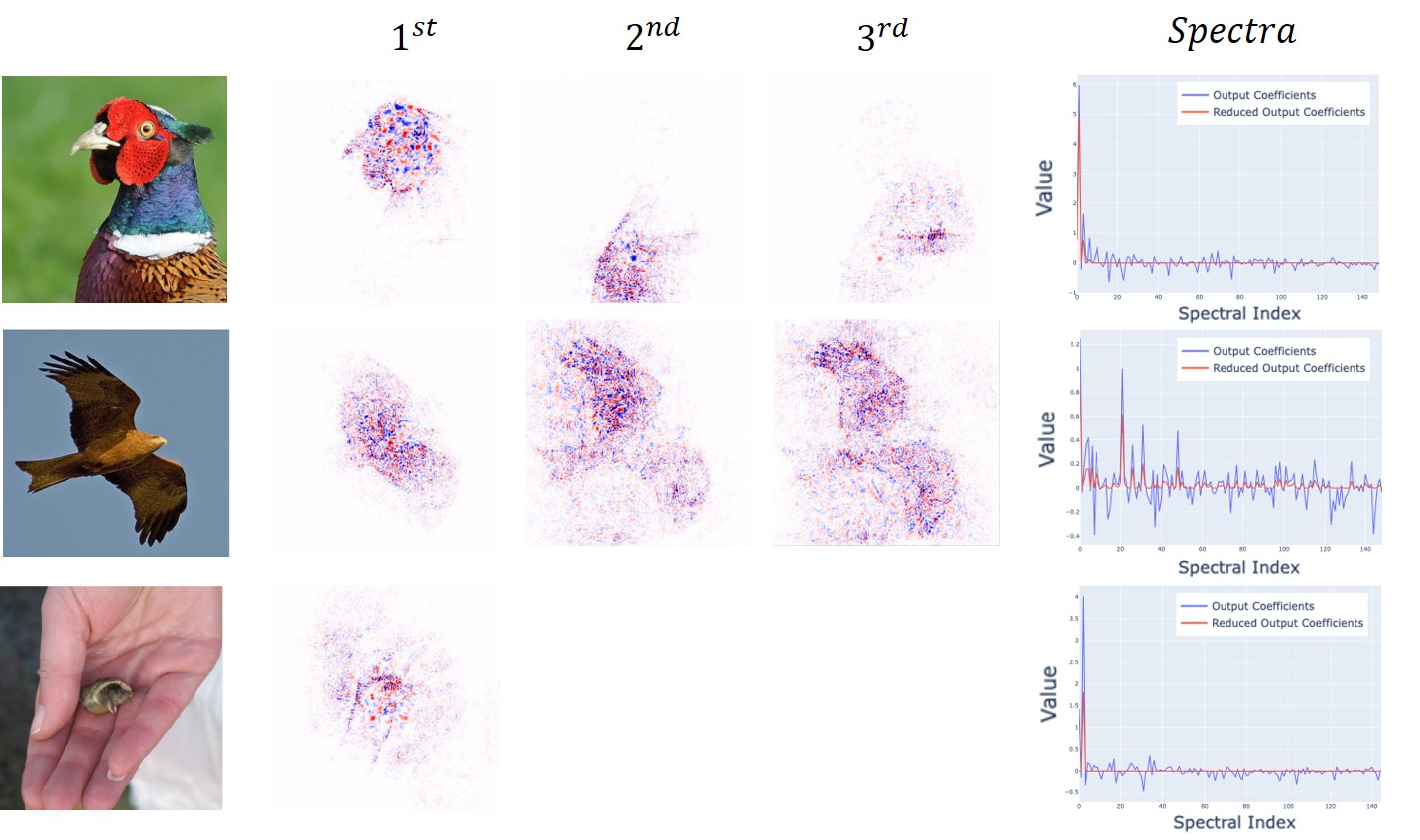}
  \caption{}
  \label{fig:o_last_conv_AN}
\end{subfigure}\hfil
\begin{subfigure}{0.45\linewidth}
  \centering
  \includegraphics[width=\linewidth]{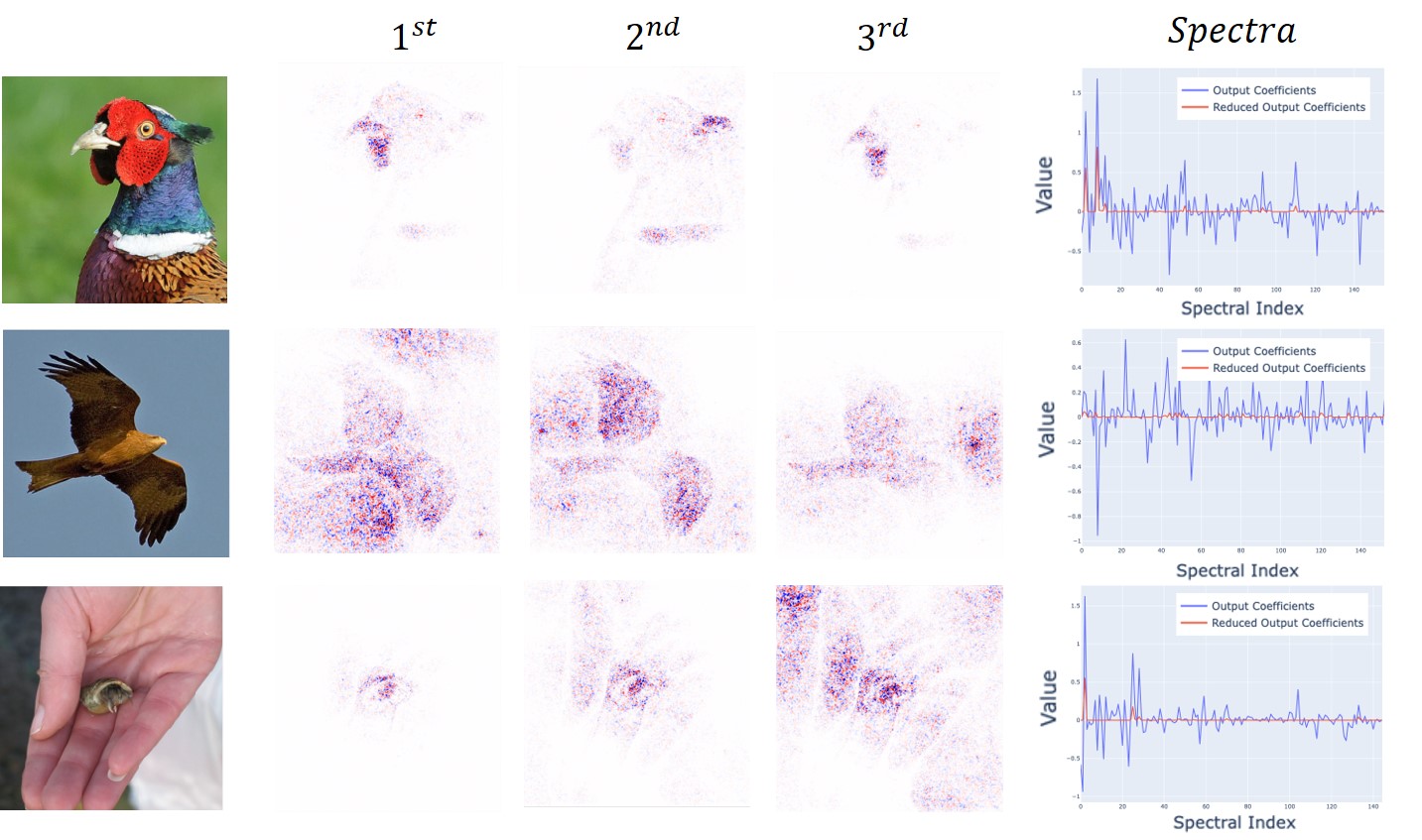}
  \caption{}
  \label{fig:o_last_conv_VGG}
\end{subfigure}
\caption{Comparison of top 3 SV's contributing to output (in descending order) and the full and reduced spectra for (a) AlexNet and (b) VGG11 from the last convolutional layer}
\label{fig:o_last_conv_top_SVs}
\end{figure}

\subsection{3D Data}

\begin{figure}[ht!]
\centering
\includegraphics[width=0.5\linewidth]{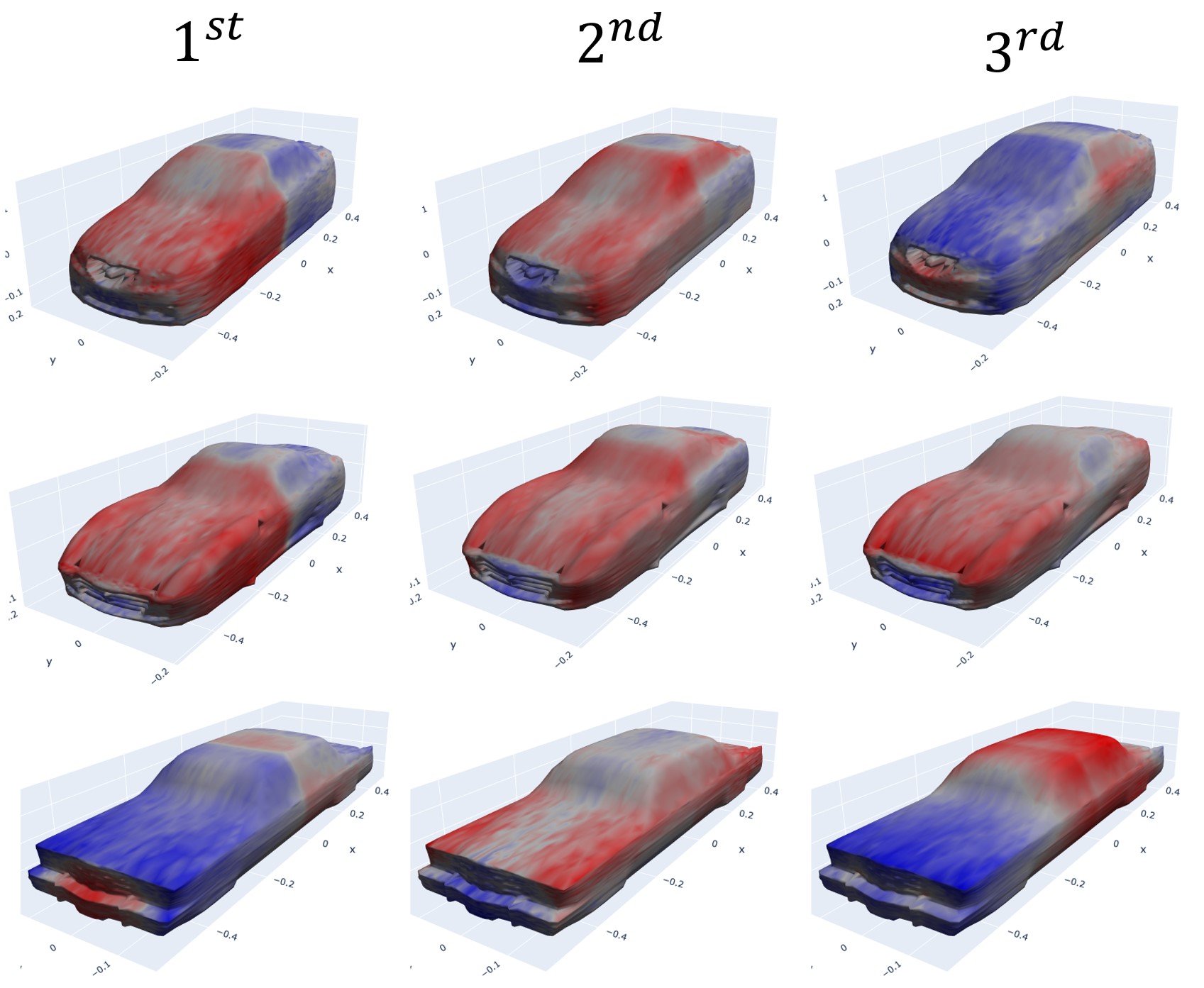}
\caption{Top 3 singular vectors influencing drag coefficient for car model}
\label{fig:car_top_3}
\end{figure}

In this section, we outline some results obtained for two 3D data sets comprising of car models and wind turbines. These are used to train a simple autoencoder whose latent space is then used to train a separate \textit{downstream} model to predict a quantity of interest. We train models to predict the drag coefficient for the cars and the peak stress for the wind turbines. The model is well-trained and provides good accuracy in prediction. In Figure \ref{fig:car_top_3}, we see the top three SV's influencing the drag coefficient of the car from the first layer of the network (in contrast with the treatment of the ImageNet models). Unlike for AlexNet and VGG the singular vectors seem more spatially coherent and seem to highlight geometrically significant parts of the car such as the front, sides, and roof.  

\begin{figure}[ht!]
\centering
\includegraphics[width=0.8\linewidth]{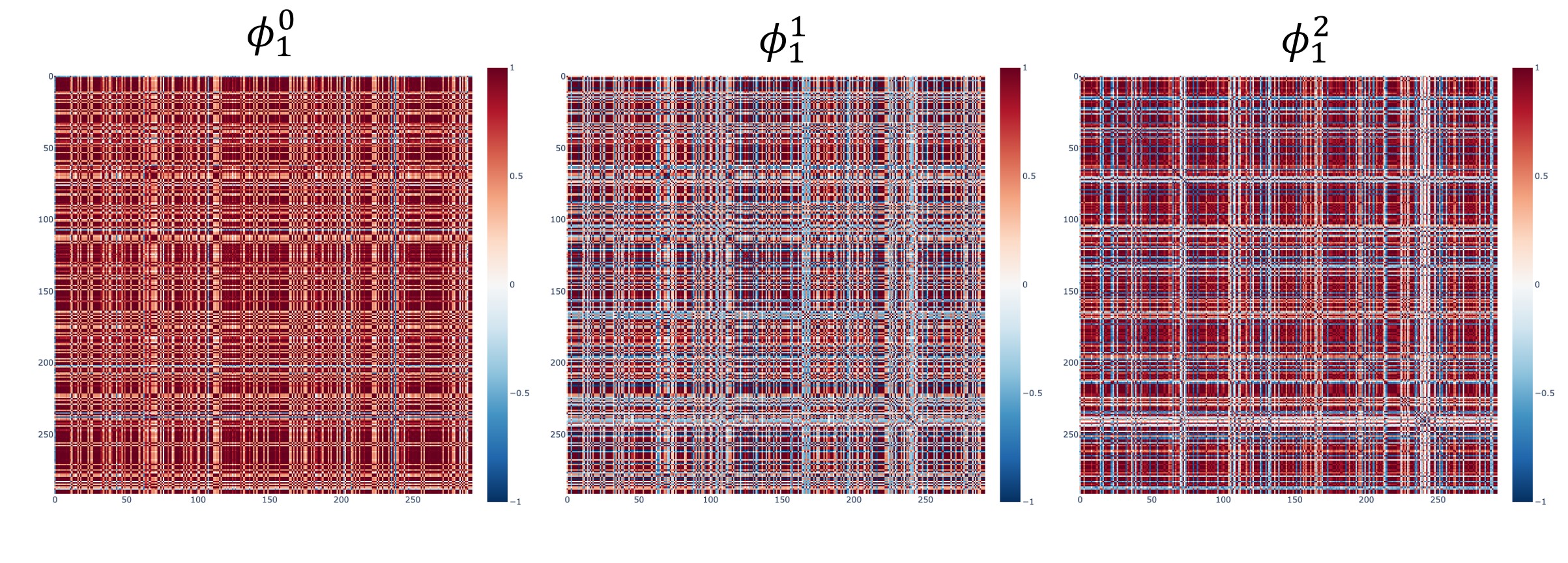}
\caption{Similarity of SV's across the car dataset}
\label{fig:phi_sims}
\end{figure}

Further, we note an apparent similarity between the singular vectors for different car shapes, suggesting that the network is making use of some generalized features, e.g. the first two SV's in the first two rows and the last SV and the first in the first and last row respectively. In order to explore this similarity we plot the inner product between SV's across the dataset in Figure \ref{fig:phi_sims}. This effectively amounts to a cosine similarity metric since the SV's are unit normalized. The first singular vector seems to be quite common across the dataset with the next two decreasing in generality but still maintaining significant subgroups. 

\begin{figure}[ht!]
\centering
\includegraphics[width=0.8\linewidth]{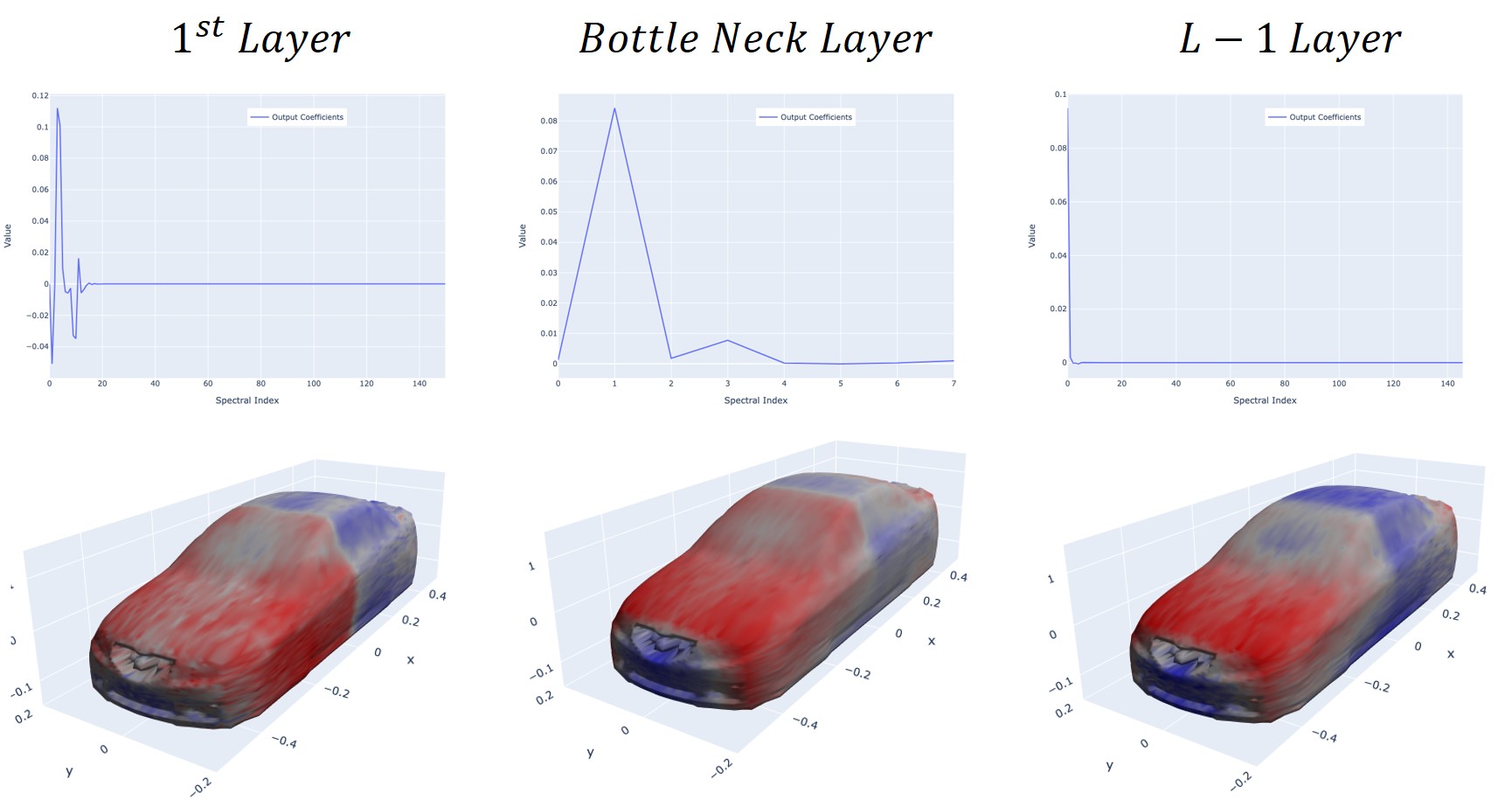}
\caption{Evolution of the top SV from the first layer on the left to the penultimate layer on the right}
\label{fig:phi_evol}
\end{figure}

We compare the evolution of singular vectors across the network by visualizing the top SV and the corresponding spectrum from the first, middle, and penultimate layers in Figure \ref{fig:phi_evol}. Two trends appear, the first being that the spectrum narrows as the depth of the layer increases with shifting contributions until the penultimate layer uses the first SV almost exclusively.  

The second trend is that the top SV seems to evolve with depth, evidenced by how certain regions of the singular vector shift the sign of their contributions and their spatial coherence increases. It is difficult to say whether this is supportive of the widely held view that models learn more abstract features with increasing depth. On the one hand, the increasing sparsity of the spectrum does suggest that features are being consolidated, yet, on the other hand, the similarity of the SV's from each layer contradicts the idea of progressive abstraction. 

\begin{figure}[ht!]
\centering
\includegraphics[width=0.8\linewidth]{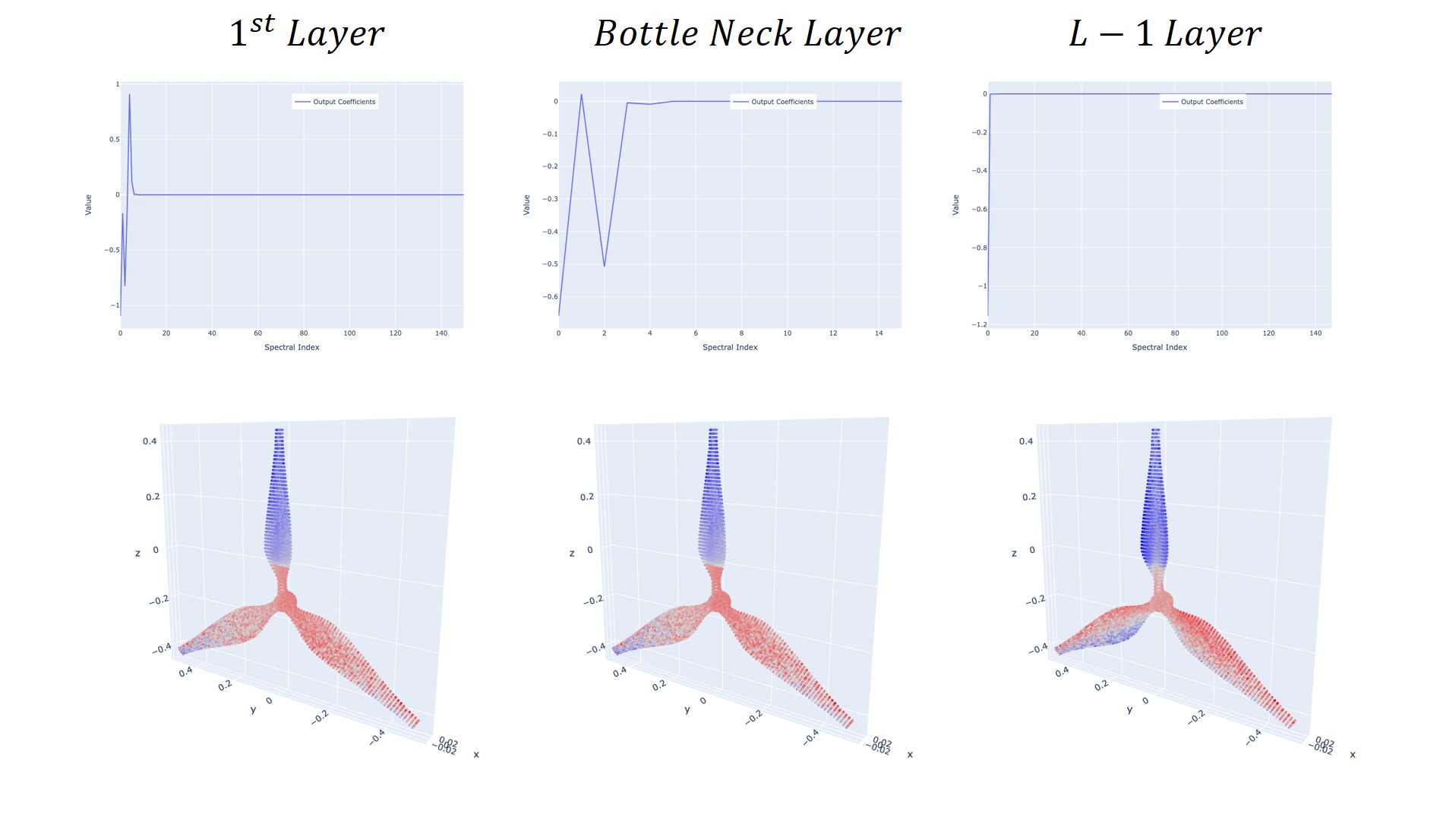}
\caption{Evolution of the top SV from the first layer on the left to the penultimate layer of the model trained on the turbine data set}
\label{fig:phi_evol_turbine}
\end{figure}

A remarkable feature of the SV's for the car data set is how they respect the symmetry of the design. It may be tempting to assume that this is always the case with symmetric designs. However, our next dataset and model were trained to reconstruct wind turbines and predict their maximum stress. In Figure \ref{fig:phi_evol_turbine}, we see that the top SV does not concentrate on the geometric features in a symmetric way. Instead, the model chose to source different contributions from each blade separately but within the same singular vector. Beyond this difference, there is a similar evolution with the depth of the SV's and the increasing sparsity of the spectra as was observed for the car dataset. 

\subsection{Auto Encoded Rotated Squares}

We end our results section by presenting our method applied to a toy dataset of randomly rotated squares of size $32 \times 32$ on a black background of size $64 \times 64$. We then train an auto-encoder network using fully connected layers \textit{without} bias, with ReLU activations and a bottleneck layer of size 8. 

\begin{figure}[ht!]
\centering
\begin{subfigure}{0.4\linewidth}
  \centering
  \includegraphics[width=\linewidth]{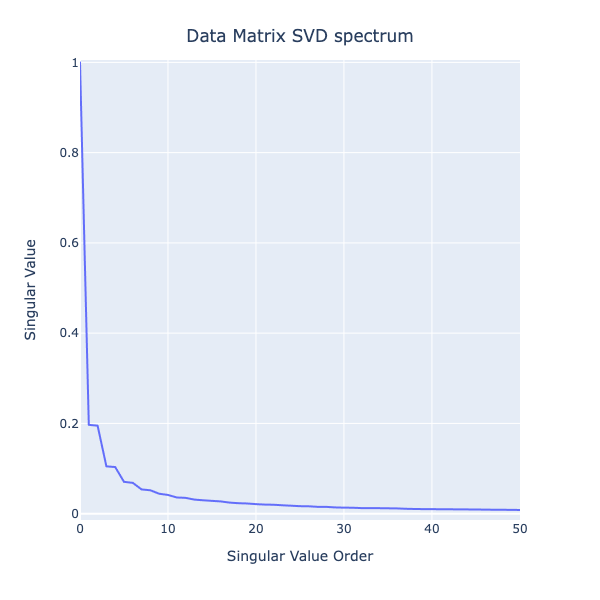}
  \caption{}
  \label{fig:data_spec}
\end{subfigure}\hfil
\begin{subfigure}{0.4\linewidth}
  \centering
  \includegraphics[width=\linewidth]{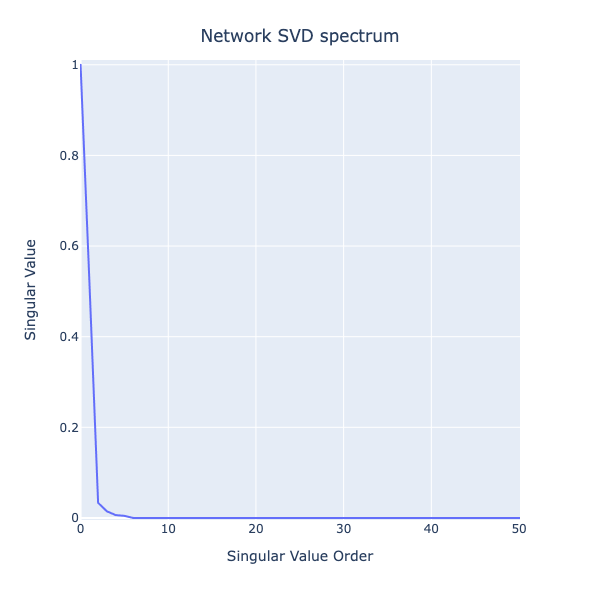}
  \caption{}
  \label{fig:net_spec}
\end{subfigure}\hfil
\medskip
\begin{subfigure}{0.6\linewidth}
  \centering
  \includegraphics[width=\linewidth]{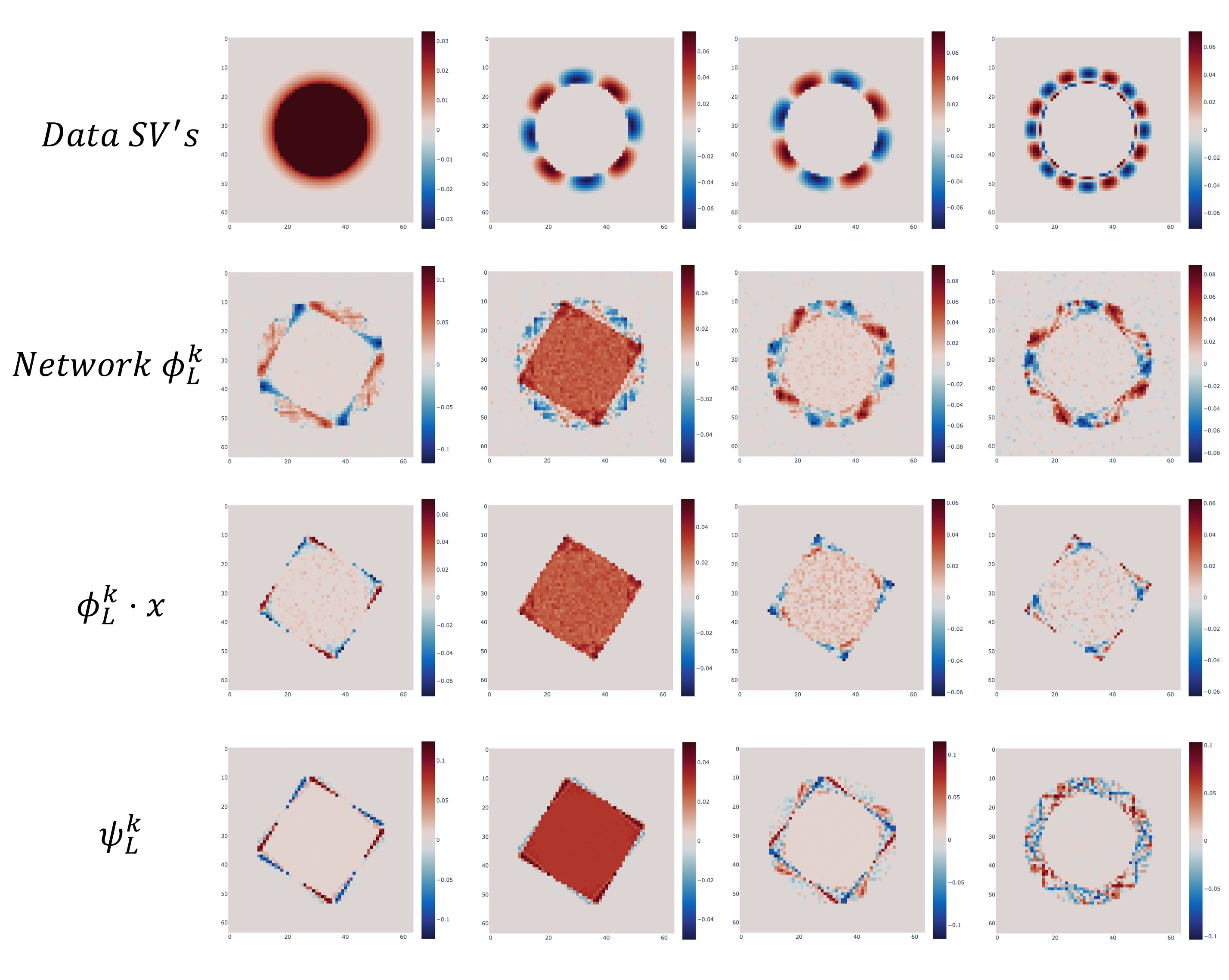}
  \caption{}
  \label{fig:data_phis}
\end{subfigure}
\caption{Comparison of singular spectra of data matrix and network linear operator. (a) Spectrum of data matrix. (b) Spectrum of network operator. (c) (1$^{st}$ row) First four singular vectors of the data matrix, (2$^{nd}$ row) Right SV's of the entire Network, (3$^{rd}$ row) Feature wise contractions of the SV's, (4$^{th}$ row) Left SV's of the entire network}
\label{fig:squares_spectra_and_phi}
\end{figure}

After training, we compare the learned singular vectors vs. the vectors obtained via SVD of the data matrix (i.e. the matrix whose rows are the flattened images). On the model side, the singular vectors are obtained by selecting random images from the test data set and calculating the SVD decomposition of the full affine operator for that instance (i.e. $l_s = L$). 

The singular value spectrum of the data matrix SVD is shown in Figure~\ref{fig:data_spec} along with the first four singular vectors in Figure~\ref{fig:data_phis}. The corresponding spectrum from the SVD decomposition of the linear operator decays much faster, suggesting that the learned linear operator is more specialized to the instance given. This view is reinforced by the visualization of the top four singular vectors. In contrast to the data-derived singular vectors that are clearly circular the network vectors are much more reminiscent of the square, although, circular aspects are also present. It is curious that the fullest square is not the top SV as the disc is for the data. Or that the other SV's are required at all since the second SV seems sufficient to reconstruct the square. However, if we consider the coefficients in (\ref{coeffs}) we see that $c^0_L = 5$ and $c^1_L = 25$ with all other contributions $c_{i>1} \ll 1$. This effectively renders $\phi^1_L$ the top contributor with edge corrections from $\phi^0_L$ as can be determined by the corresponding SVs of the $U_L$. We note that the singular vectors of the $U_L$ and $V_L$ matrices are complementary but have significant differences. 

\section{Discussion}

\subsection{Bias term contribution}

So far we have concentrated on understanding the $u_i$ operator through its singular vectors and spectrum. However, it is not the only input-dependent term contributing to the output and here we aim to highlight that the bias term has a significant role to play. 

\begin{figure}[ht!]
\centering
\subcaptionbox{\label{fig:wnb}}{\includegraphics[width=0.6\linewidth]{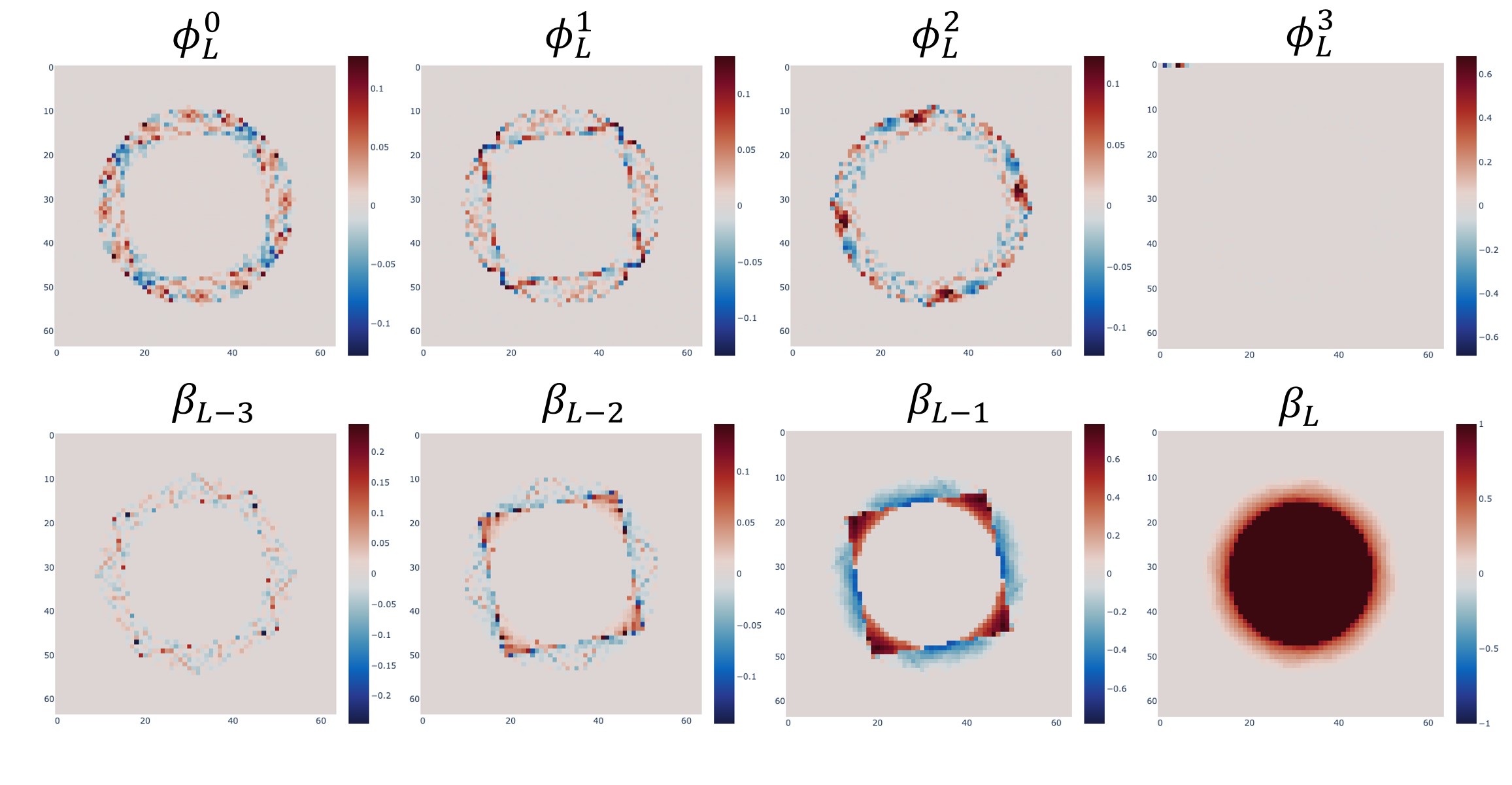}}\hspace{1em}%
\subcaptionbox{\label{fig:recs}}{\includegraphics[width=0.16\linewidth]{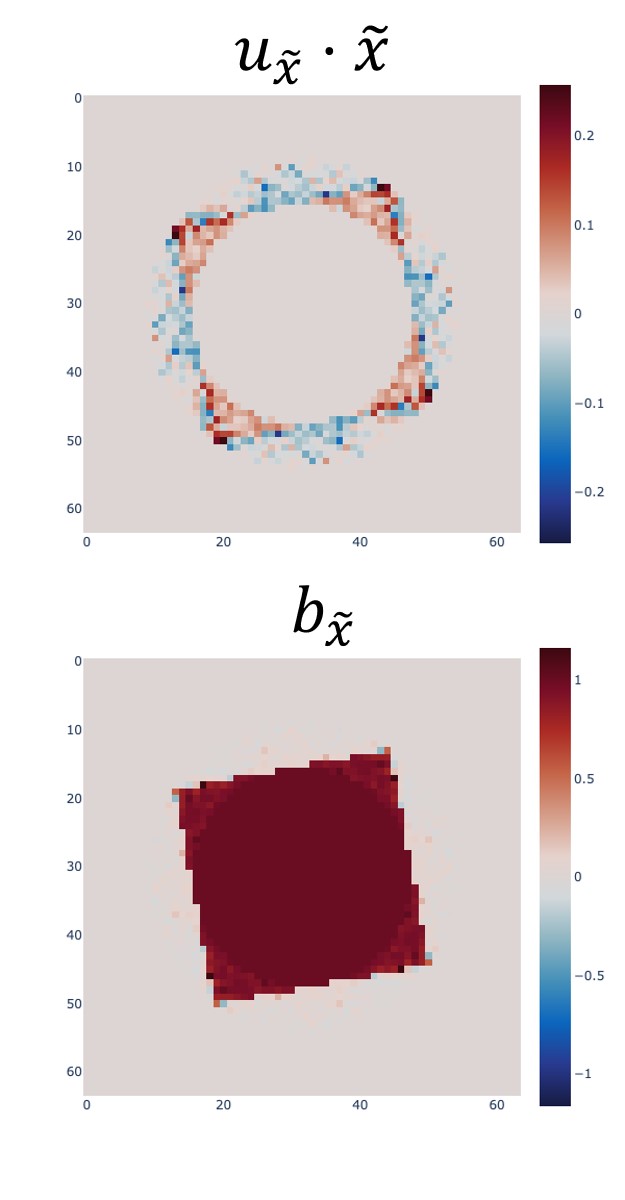}}%
\caption{Comparison of contributions coming from the SV's of the $u_{\tilde{x}}$ term and those coming from the bias terms. (a) (top) Top 4 singular vectors, (bottom) bias term contributions from the decoder layers. (b) (top) $u_{\tilde{x}} \cdot \tilde{x}$, (bottom) $b_{\tilde{x}}$ as calculated from the RHS of (\ref{bias_term})}
\label{fig:weights_and_biases}
\end{figure}

For the toy problem, we trained a network without any bias terms in order to make the comparison with data SVD. We train a new network with bias terms and explore their contributions. Remarkably, the SV's of the new network only correct the edges of the bulk that is generated by the bias terms as shown in Figure \ref{fig:weights_and_biases}. This is quite different from the solution found by the no-bias network and is, somewhat, counter-intuitive though the bias term from the last decoder layer does align nicely with the top SV from the data matrix.

\begin{figure}[ht!]
\centering
\subcaptionbox{}{\includegraphics[width=0.6\linewidth]{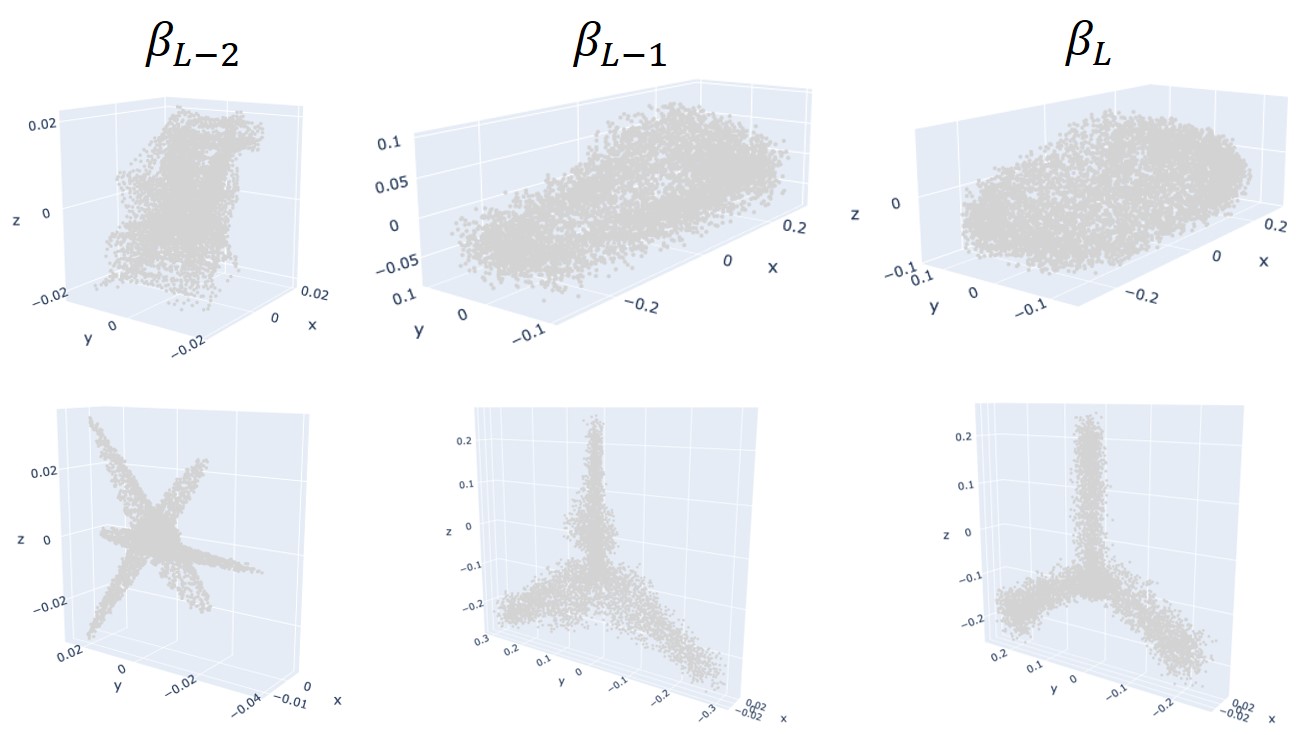}}\hspace{1em}%
\subcaptionbox{}{\includegraphics[width=0.23\linewidth]{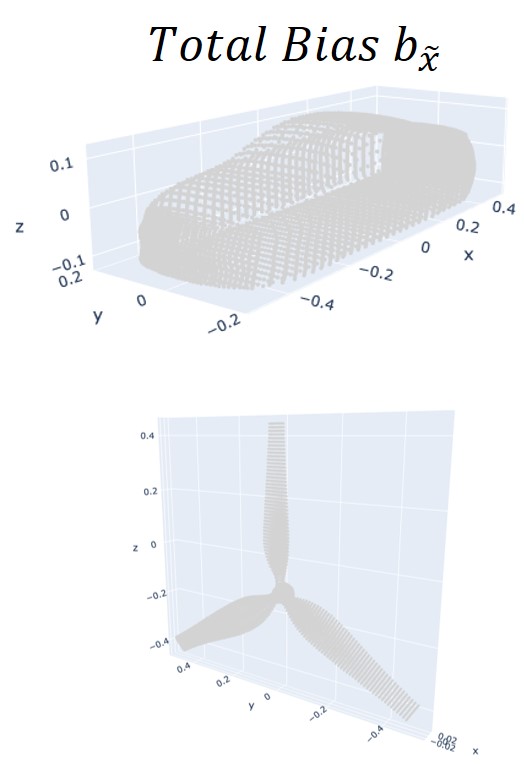}}%
\caption{Bias contributions of (a) the last three layers and (b) the total bias term}
\label{fig:3d_biases}
\end{figure}

We can apply a similar analysis for the 3D data since we trained an autoencoder to reproduce the input shape before the downstream model. In Figure~\ref{fig:3d_biases} we show the bias contributions of the last few layers of the autoencoder. 

Similar to the toy data the bias terms of the last two layers are responsible for the bulk of the reconstructed shape. An intriguing difference is how the bias contributions look like noisy point clouds but their sum results in a highly structured shape. 

It seems that the bias terms can also be responsible for the representative power of deep models and have been quite overlooked in the XAI literature. A possible reason for this behavior and the increased performance over the no-bias network might be due to the ease of training the bias terms through their shallower gradient paths. This would be in line with the motivations and performance of ResNet~\citep{he2016deep} architectures.

\subsection{Feature wise contraction}

\begin{figure}[h!]
\centering
\includegraphics[width=0.8\linewidth]{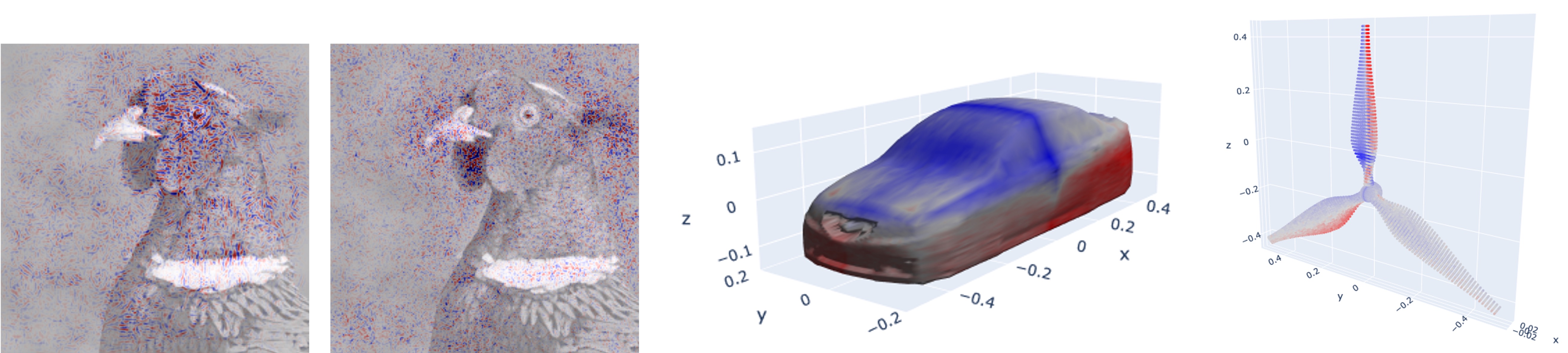}
\caption{Feature wise averaged top contributing SV's for AlexNet, VGG11, cars and wind turbines}
\label{fig:non_contracted}
\end{figure}

Throughout the results section we have made use of the feature-wise contraction to generate the heatmaps displayed. However, it may not be clear why this is more interpretable than using the SV's directly, especially since for the toy dataset the SV's don't look too far from their contracted versions. We display some examples from the image and 3d data sets in Figure \ref{fig:non_contracted} where the feature channels are averaged over instead of contracted with the input. 
The most distinct difference between the two treatments for the images is how much more diffuse and present on the background the averaged SV's are. For the 3D models, the SV's are not as diffuse but they do give a false impression of the contribution each point brings to the output. This is also the case for the images where the averaged SV's may lead us to over-attribute the importance of the background, when in fact it gets canceled pixel-wise. This pixel/point-wise cancellation without taking the corresponding weights to zero is a remarkable feature of the learned operators. It reinforces the view that learning to ignore features is just as important part of learning as paying attention to the \textit{main} signal. 

\subsection{Comparison with other XAI methods}

\begin{figure}[h!]
\centering
\includegraphics[width=0.8\linewidth]{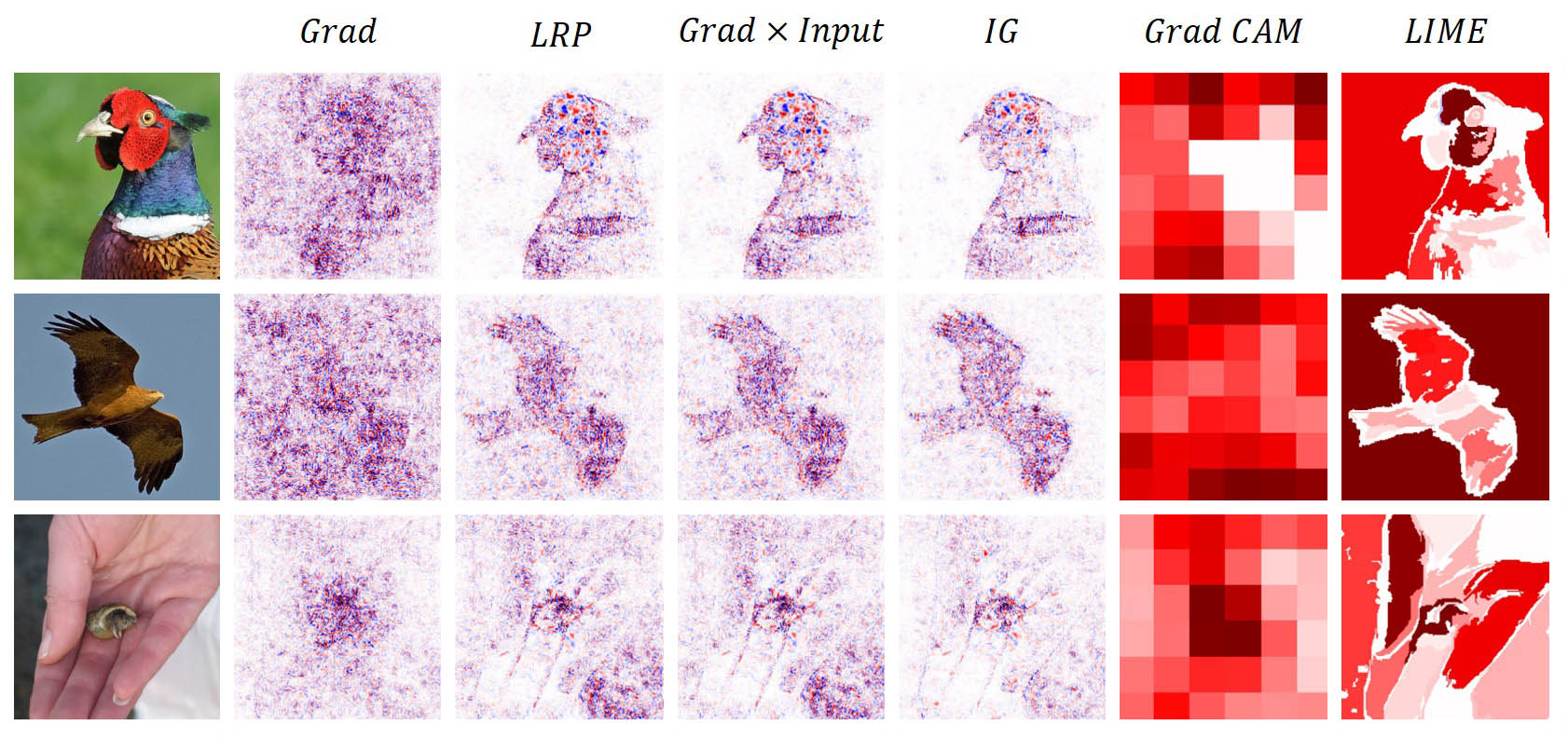}
\caption{Comparison explanations for AlexNet}
\label{fig:alex_net_comp}
\end{figure}

\begin{figure}[ht!]
\centering
\includegraphics[width=0.8\linewidth]{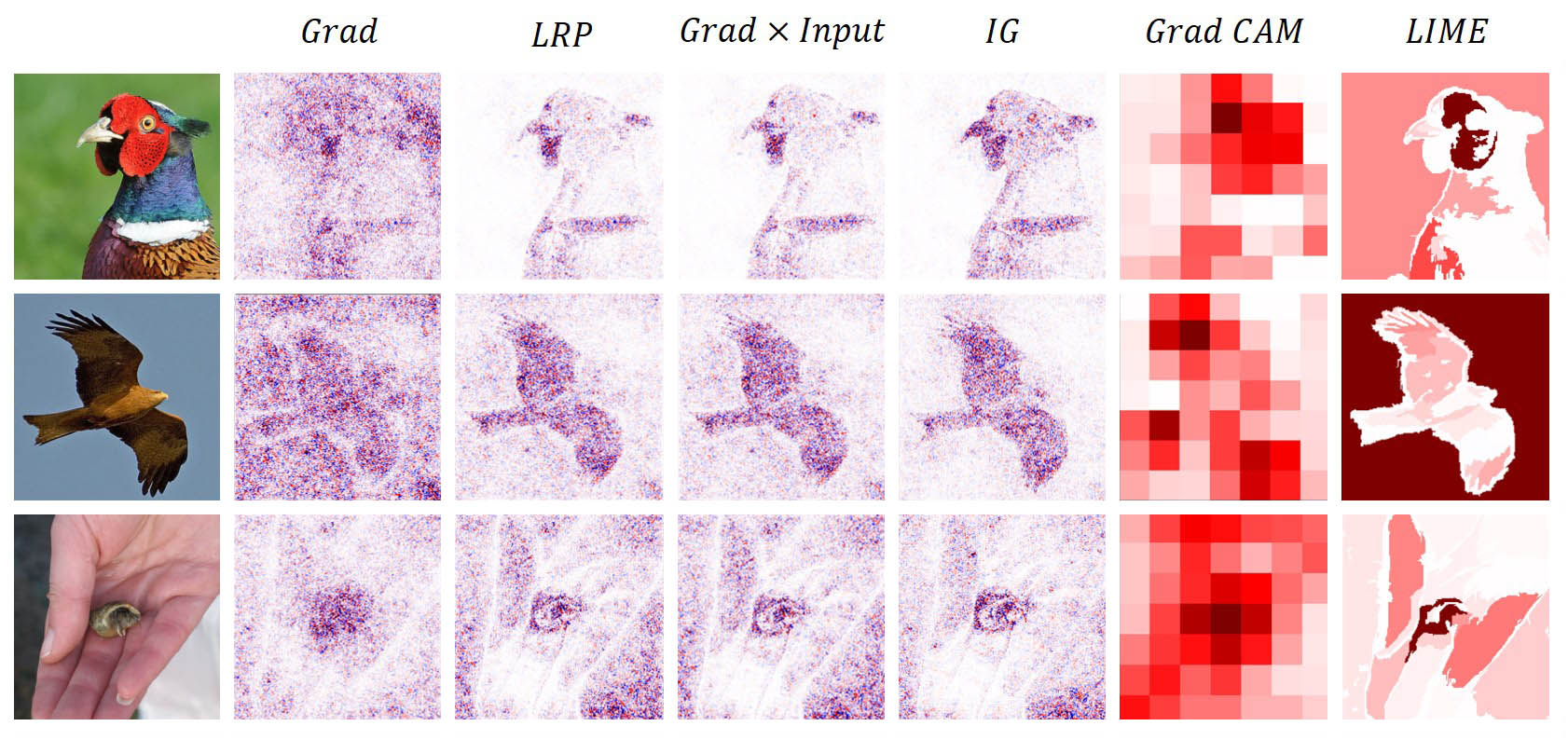}
\caption{Comparison explanations for VGG 11}
\label{fig:vgg_comp}
\end{figure}

For comparison, we present here explanations produced by a prominent subset of the many XAI methods available in the literature. Since most of these methods were developed with 2D computer vision in mind we restrict our comparison to the ImageNet models. In Figures \ref{fig:alex_net_comp} and \ref{fig:vgg_comp} we see attribution heatmaps for the three ImageNet examples used throughout our results. 

\textbf{Gradient based \cite{Simonyan2013,shrikumar2016just,sundararajan2017axiomatic} and LRP \cite{binder2016layer}} explanations seem in good agreement with each other. Notably integrated gradients (IG) \cite{sundararajan2017axiomatic} have the least amount of noise in their heatmaps while grad exhibits the most. This of course can be attributed to the feature-wise dot-product and averaging effect of the integral in IG. 

There is a lot of similarity between these explanations and the singular vectors produced by the penultimate layer. This can be expected since the right operator in (\ref{left_right}) is most of the network. However, with the existing XAI methods, there is no opportunity to decompose the effect of the network into potentially less noisy pieces. The most striking difference is that we can naturally inspect the operators and representation at an intermediate layer and compare SV's in an equivalent way. In doing so, the advantage of our method becomes most apparent and we can see how the operator coming from the last convolutional layer has SV's that are spatially separated.  

\textbf{GradCam \cite{Selvaraju_2019}} also attempts to understand the network in terms of the output of the last convolutional layer. However, due to the averaging operations and the necessary interpolation, the produced explanations are very coarse. In contrast, our method is fine-grained, and the SV's highlight how the filters decompose the input. 

\textbf{LIME \cite{Ribeiro2016}} builds a linear surrogate of the model around perturbations of the input applied to superpixels obtained by Felzenszwalb segmentation \cite{felzenszwalb2004efficient}. Despite the fact, the models considered are already linear around the input the produced explanations are quite different than the SV's and other gradient-based methods. The biggest disagreement is the over-importance LIME assigns to the background super pixel. This may be explained by either the fact that Lime produces out-of-distribution perturbations or because the background gets removed via the feature-wise dot product in the other methods including our own.   

\section{Method}
\label{sec:method}


In this section, we outline the mathematical preliminaries underpinning our method. 

\subsection{Deep networks}

A deep neural network, in the most general sense, is a map $f_{\theta}: \mathbb{R}^M \to \mathbb{R}^N$ between an input signal $x \in \mathbb{R}^M$ to the space of desired output signals $y \in \mathbb{R}^N$, where $\theta$ denotes all parameters of the map. It is important to note that regardless of the actual input and output signal type (e.g. binary or categorical) deep learning usually assumes the underlying algebraic field is continuous. The term \textit{deep} refers to the fact that the map $f_{\theta}$ is a composition of multiple maps, that are loosely termed \textit{layers}, each with their own set of parameters 

\begin{equation} \label{composition}
    f_{\theta}(x)  = (f_{\theta_L}^L \circ f_{\theta_{L-1}}^{L-1} \circ \dots  \circ f_{\theta_{1}}^{1})(x)  \qquad \theta_i \subset \theta
\end{equation}

where $L \in \mathbb{N} >2$ specifies the number of layers. The prototypical layer consists of an affine map followed by an element-wise non-linear \textit{activation function} $\sigma$. 

\begin{equation} \label{rep}
    z_{l+1} = \sigma(W_{l}z_l+b_l)
\end{equation}

Where $z_0 = x$, $z_L = y$ and the intermediate cases $z_l$ for $1<l<L $ are termed hidden or latent representations. The non-linearity $\sigma$ is usually chosen from a popular set of functions such as $\text{Sigmoid}$, $\text{Tanh}$ or $\text{ReLU} := max(\cdot,0)$ and their variants. These non-linear activations are responsible for the ability of neural networks to approximate arbitrary functions and are the source of the \textit{observed} representational power of deep learning. Indeed, without the non-linearity the composition of multiple linear maps $f_{W,b}(x) = Wx+b$ would collapse into a single linear transformation. 

\begin{equation} \label{colapse}
    f_{W_2,b_2}\left(f_{W_1,b_1}(x)\right) = f_{W_2W_1,W_2b_1+b_2}
\end{equation}

Due to the combination of a large number of layers, a large number of parameters, and the use of non-linear activation functions, deep learning models have come to be perceived as black boxes. 

\subsection{Local Linearity}

Despite the global non-linear nature of deep neural networks, it was noticed in \citet{montfar2014number} that certain architectures containing piece-wise linear activation functions are completely defined by the separate linear pieces that it's composed of. More specifically, if we restrict ourselves to a linear region of the input domain $ R_{\ihat} \subset \mathbb{R}^M$ then for any input in that region the network output is given by: 

\begin{equation} \label{lin_region}
    y_{\ihat} = u_{\ihat}\cdot x_{\ihat} + b_{\ihat}, \qquad x_{\ihat} \in R_{\ihat}
\end{equation}

This is known as the input-dependent point-wise affine map (PWA) form of the network. The above equation shows that in a linear region, the network is reduced to a matched filter output plus a bias term or, put another way, a dot product between the input and a learned vector. For the case when the network is composed of linear transformations and $\text{ReLU}$ activation functions we can explicitly write the vector as

\begin{equation} \label{u_vec}
    (u_{\ihat})_j = (W_L)_{j:}\cdot \text{diag}(I_{z_{L-1}>0})\cdot W_{L-1}\cdot \, \dots \, \cdot \text{diag}(I_{z_{1}>0})\cdot W_{1}
\end{equation}

Where the operator $\text{diag}(I_{z_{l}>0})$ is represented by a diagonal matrix of element-wise indicator functions. By noticing that the indicator operator is in fact the derivative of the $\text{ReLU}$ activation function we can write the following:

\begin{equation} \label{d_layer}
    \frac{\partial}{\partial z} \text{ReLU}(f_{W,b}(z))  = \frac{\partial \text{ReLU}(z)}{\partial z}\Big|_{f_{W,b}(z)}\cdot\frac{\partial}{\partial z}  (f_{W,b}(z)) = \text{diag}(I_{f_{W,b}(z)>0}) \cdot W
\end{equation}

As previously noted deep networks are compositions of differentiable functions and we can exploit the chain rule to efficiently calculate the $u_i$ vector.

\begin{equation} \label{u_vec_1}
    (u_{\ihat})_j = \left(\frac{\partial z_L}{\partial z_{L-1}}\right)_{j:} \cdot \frac{\partial z_{L-1}}{\partial z_{L-2}}\cdot \, \dots \, \cdot \frac{\partial z_{1}}{\partial z_0} = \left(\frac{\partial y}{\partial x}\right)_{j:}
\end{equation}

In practice, we do not know apriori, what the linear regions are. Fortunately, we can still obtain the corresponding linear transformation by differentiating the network w.r.t. a specific example of interest $\tilde{x}$ in the input space.

\begin{equation} \label{linear_out}
    f_{\theta}(\tilde{x}) = \frac{\partial f_{\theta}(x)}{\partial x} \Big |_{\tilde{x}} \cdot \tilde{x} + b_{\tilde{x}} = u_{\tilde{x}} \cdot \tilde{x} + b_{\tilde{x}}
\end{equation}

The bias term can be obtained in two ways:

\begin{equation} \label{bias_term}
    b_{\tilde{x}} = f_{\theta}(\tilde{x})-\frac{\partial f_{\theta}(x)}{\partial x} \Big |_{\tilde{x}} \cdot \tilde{x} = \sum_{l=1}^L\frac{\partial f_{\theta}(x)}{\partial b_l} \Big |_{\tilde{x}} \cdot b_l = \sum_{l=1}^L \beta_l
\end{equation}

Now we can obtain the linear regions by perturbing $\tilde {x}$ and testing for change in the operator. For this particular architecture, the input dependence is only manifest through the indicator functions since the other terms correspond to the layer weights. This means that two inputs with the same activation signs will result in the same affine map and will belong to the same linear region $R_i$.    

We can view the model as an ensemble of linear models with restricted support to the linear regions. Since linear models are considered inherently interpretable, then a neural network with piecewise activation functions can be said to be \textit{effectively} interpretable when considering its form for a specific input.  

\subsection{The class of locally linear models}

So far we have shown local linearity for networks consisting of affine maps followed by ReLU activation functions. However, the class of locally linear models is substantially broader as we show in the following.

\textbf{Convolutional layers} are a popular choice for vision-related tasks and signal processing. The inputs to these layers are usually organized as tensors with spatiotemporal dimensions and channel (features) dimensions e.g. for video $z^{t\, h\, w\, c} \in \mathbb{R}^{T} \otimes \mathbb{R}^{H} \otimes \mathbb{R}^{W} \otimes \mathbb{R}^{C}$. These inputs are then convoluted with the layer filters $W^{t\, h\, w\, c\, c'} \in \mathbb{R}^{T_k} \otimes \mathbb{R}^{H_k} \otimes \mathbb{R}^{W_k} \otimes \mathbb{R}^{C}\otimes \mathbb{R}^{C'}$

\begin{equation} \label{conv}
\tilde{z}^{t'\, h'\, w'\, c'} = \sum_{\tau = 0}^{T_k-1}\sum_{\eta = 0}^{H_k-1}\sum_{\omega = 0}^{W_k-1}\sum_{\kappa = 0}^{C_k-1} W^{\tau \,\eta \,\omega \, \kappa\, c'} z^{\tau+t'\, \eta+h'\, \omega+w'\,  \kappa} +b^{c'}
\end{equation}

Despite the above succinct representation the most efficient way to implement convolutions is to flatten the input into a vector and matrecize the weights into a Toeplitz matrix. Ultimately, convolutions are affine maps with a constrained weight matrix $\tilde{W} \in \mathbb{R}^{T'H'W'C'\times TWHC}$ 

\begin{equation} \label{flat_conv}
    \tilde{z} = \tilde{W} \cdot z + \tilde{b} \qquad \tilde{z},\tilde{b} \in \mathbb{R}^{T'H'W'C'} ;\, z \in \mathbb{R}^{THWC}
\end{equation}

The output dimensions are a function of the convolution kernel size, dilation, stride, and padding.  

\textbf{Pooling layers} are a type of convolution operation aimed at reducing the size of the spatiotemporal dimensions while leaving the channel dimension the same. The weights of these layers can be fixed as in the case of average pooling or can be input dependent like max pooling, however, since they are convolution layers they too are effectively affine maps.

\textbf{Residual layers} a.k.a. skip connections, are an effective way to combat the vanishing gradient problem by introducing an additive operation and allowing paths of a lower depth. 

\begin{equation}
    y = (W + f) \cdot x + b
\end{equation}

Where $f$ is a locally linear deep network and $W$ is a map, not necessarily trainable, used to match the input to the same dimension as the image of $f$. Residual layers are pointwise affine as long as $f$ is pointwise affine. 

\textbf{Concatenation layers} are a method of combining inputs from different branches into a single representation. Taking $z_1 = f_1 \cdot x$ and $z_2 = f_2 \cdot x$ to be the outputs of two locally linear branches of the network, a concatenated layer can be expressed as 

\begin{equation}
    z = W\cdot [z_1\,z_2] + b = [W_1 \,W_2] \cdot [z_1\,z_2] + b = W_1\cdot z_1 + W_2 \cdot z_1 +b = (W_1f_1+W_2f_2)\cdot x +b
\end{equation}

Which is equivalent in form to a residual layer and hence pointwise affine.

\textbf{Activation functions} have already been identified as element-wise non-linear operations critical to model performance. However, despite their non-linearity they too admit an input dependent point wise affine representation 

\begin{equation}
    \sigma(z) = f_{\Lambda(z),b}(z) = \text{diag}\left(\frac{\sigma(z) - b}{z}\right)\cdot z + b     
\end{equation}

Note that this representation is not unique, and we have the freedom to choose any constant $b$ including $0$ or $\sigma(0)$. For the special case, when the activation is piecewise linear, the canonical representation is given by the gradient of the activation function as was shown for the ReLU case above.

\begin{equation}
    \Lambda(z) = \text{diag}\left(\frac{\partial\sigma(z)}{\partial z}\right)   
\end{equation}

This representation is also valid for the near linear regions of Sigmoid and Tanh activations and is favored in this work due to its ease of computation using auto differentiation packages. 

We define the class of locally linear models as functions formed by composing successive pointwise affine maps. Their local linearity can be proven by noticing that these maps map points to points and that their composition is again pointwise affine.  As demonstrated above, many popular components of deep learning models fall into this category and the class of locally linear models is quite broad. 

\subsection{Spectral surgery}

Having shown that a large class of deep learning models can be reduced to locally linear models we now turn to the analysis of these models for the purposes of explainability. The natural question is how do we analyze the overall effect of composing multiple pointwise affine maps. One method is to focus on the final output and analyze the effective linear transformation $u_i$ in (\ref{lin_region}). Indeed, this strategy is used for saliency maps where the gradient of the output with respect to the input gives $u_i$ as shown in (\ref{u_vec_1}). 

However, despite it being an exact representation of the network action, it is an aggregated view that gives limited insight into the various components that make up the final prediction. The main contribution of this work is to decompose the output of the model into a linear combination of spectral components that are ranked by their contribution. 

The first step is to pick an intermediate layer and separate the chain of pointwise affine maps (\ref{u_vec_1}) into a left and right piece.
\begin{equation} \label{left_right}
    (u_{\tilde{x}})_j = \left(\prod_{l=l_s}^L W_l \right)_{j:}\left(\prod_{l=1}^{l_s} W_l \right)\cdot \tilde{x} = (L_{l_s})_{j:}R_{l_s}\cdot \tilde{x}   
\end{equation}

Where we have used $W_l$ for the representation of all pointwise affine maps involved. Next, we obtain the SVD of the right operator 

\begin{equation}
     L_{l_s}R_{l_s}\cdot \tilde{x} = L_{l_s}U_{l_s}\Sigma_{l_s}V^T_{l_s}\cdot \tilde{x} = \hat{L}_{l_s} \cdot c_{l_s}
\end{equation}

Where $\hat{L}_{l_s} = L_{l_s}U_{l_s}$ is a spectral left operator and $ c_{l_s} = \Sigma_{l_s}V^T_{l_s}\cdot \tilde{x}$ are spectral coefficients obtained by projecting the input onto the singular orthonormal vectors $\phi^i$ which are the columns of $V_{l_s}$ and scaling by the corresponding singular values $\lambda^i$

\begin{equation} \label{coeffs}
    c_{l_s}^i = \lambda^i_{l_s} \phi^i_{l_s} \cdot \tilde{x} \qquad \phi^i_{l_s} \cdot \phi^j_{l_s} = \delta^{ij}
\end{equation}

Where $\delta^{ij}$ is the Kronecker delta function. The singular values are defined to be positive and ordered in descending order $\lambda_0 > \lambda_1 > \dots \lambda_{r-1}$ where $r$ is the rank of $L_{l_s}$. This decomposition of the input into the singular vectors (SV) of the operator is in contrast with the usual PCA methods used in statistics and data science but in line with the spectral methods common in physics and the analysis of differential equations. Specifically, given a linear differential operator $L$ and problems of the form

\begin{equation}
    Lu = f
\end{equation}

The solutions are obtained by expanding both $u$ and $f$ in terms of the eigenvectors of $L$ and solving the resulting algebraic system.
We interpret the singular vectors as a set $\{\phi^i\}_{i=0}^{r-1}$ of learned high dimensional filters which can be visualized for the purposes of explainability. In addition, we can obtain a linear decomposition of the network output in terms of the excitation of these filters.

\begin{equation}
   y_{\tilde{x}} = \sum_{k=0}^{r-1} \alpha^k_{l_s} + b_{\tilde{x}} = \sum_{k=0}^{r-1} \psi^k_{l_s} \lambda^k_{l_s} \phi^k_{l_s} \cdot x_{\tilde{x}} + b_{\tilde{x}}         
\end{equation}

Where the $\psi^k_{l_s}$ are the $r$ components of the row of the spectral left operator responsible for the output $(\hat{L}_{l_s})_{j:}$. It is important to note that the sign of the $\alpha^k$ may not be positive nor does the magnitude preserve the order established by the singular values. However, what we do have for certain is an additive representation of the network output in terms of likely far fewer components than the input dimension $r<<M$. 

We can further reduce the number of components by imposing a threshold either on the absolute values or the quantiles of the $\alpha^k$. Here we consider an alternative approach in which we reduce the components to purely positive or negative depending on the overall sign of the sum. To do so we separate the components into two complementary ordered lists $\alpha^+ = \{\alpha^k_j| \alpha^k_j >0\}$ and $\alpha^-$, where $j$ denotes the order index and $k$ is the spectral index. Then we add the two sequences order-wise to obtain a new set of coefficients

\begin{equation} \label{ahat}
    \hat{\alpha} =  \left(\alpha^+_k+\alpha^-_k\right)_{j = 0}^{\min(|\alpha^+|,|\alpha^-|)-1} \bigcup\, (\alpha^+_j)_{j =\min(|\alpha^+|,|\alpha^-|)}^{|\alpha^+|}\bigcup \,(\alpha^-_j)_{j =\min(|\alpha^+|,|\alpha^-|)}^{|\alpha^-|}
\end{equation}

If the new sequence still has mixed signs then we can repeat the process with the resulting sequence. Once all terms have the same sign, they can be normalized to give the proportional contribution of each singular vector

\begin{equation} \label{reduced_a}
   \tilde{\alpha}^k = \frac{\hat{\alpha}^k}{\sum_{j = 0}^{|\hat{\alpha}|}\hat{\alpha}_j} \qquad \hat{\alpha}^k,\hat{\alpha}_j \in \hat{\alpha}
\end{equation}

The justification for this procedure comes from the observation that weights in deep learning are initialized randomly with both positive and negative values while the common classification task requires either a positive or negative final output. Therefore, in the absence of a least energy constraint, solutions of many excitations that average out to a sufficiently positive or negative outcome are just as valid as ones of a single excitation.  
Note that we picked the layer $l_s$ to split the operator arbitrarily and each choice presents a new basis. However, each basis results in the same operator but expressed in terms of spaces with differing ranks. 

The spectral surgery referred to in the title of this subsection is the combination of exploring the basis of singular vectors at each layer and collapsing or pruning contributions to obtain an interpretable representation of the model.

\subsection{Feature wise contraction}

The singular vectors have the same dimensions as the input and their action on the input is that of a dot product. However, when the data has a more natural tensor representation, such as an image, then we find that it proves insightful to perform the dot product along a particular dimension. 

\begin{equation} \label{feat_contraction}
    c^k_{hw} = \sum_c \left(\phi^k_{l_s}\right)_{hwc} x_{hwc}
\end{equation}

This has a couple of advantages beginning with the fact that it is a more natural way to treat multi-dimensional attributions for each point in the input than averaging or taking absolute values. In this way, we consider the attributions as a contribution each point brings to the output, which can be zero. This is the second advantage of the contraction in that it allows us to determine which points don't affect the prediction and may emphasize the contributing points better. 

We note the similarity with the gradient times input XAI method \cite{shrikumar2016just}. However, we offer a more principled prescription to aggregate the contributions at each point compared to the usual masking of negative values. This does not permit the cancellation of features and may present an inaccurate view of the network.     

\subsection{Symbolic representation}

At this point we have all the necessary ingredients to produce a symbolic representation of the network. This is usually done in spectral decomposition methods as a change of basis.

\begin{equation} \label{change_of_basis}
    \tilde{x} = \sum_k \left( \phi^k_{l_s} \cdot \tilde{x} \right) \, \phi^k_{l_s}
\end{equation} 

In practice we find that the projections of the input onto the SV's to be of similar magnitude across the entire spectrum and thus unsuitable for our XAI purposes. This representation does not directly take into consideration how the input influences the network output. The combination of the singular values $\lambda_{l_s}^k$ and the left operator $L_{l_s}$ typically induce a natural ranking among the SV's. Therefore, we symbolically decompose the output of the network as: 

\begin{equation}
   \left(y_{\tilde{x}}\right)_{hw} = \sum_{k=0}^{r-1} \alpha^k_{l_s}\hat{c}^k_{hw} + \left(b_{\tilde{x}}\right)_{hw} \quad \Rightarrow \quad y_{\tilde{x}} = \sum_{hw}\left(y_{\tilde{x}}\right)_{hw}
\end{equation}

Where

\begin{equation} \label{feat_contraction}
    \hat{c}^k_{hw} = \frac{c^k_{hw}}{\sum_{hw}c^k_{hw}}, \qquad \left(b_{\tilde{x}}\right)_{hw} = \frac{b_{\tilde{x}}}{|h||w|}
\end{equation}

This decomposition takes account of the ranking induced by the output and incorporates the feature wise contraction. In addition we can make use of the reduction strategy to replace the $\alpha^k_{l_s}$ by $\tilde{\alpha}^k_{l_s}$ from (\ref{reduced_a}) and obtain coefficients of homogenous sign. We emphasise that although the indices reflect an image classification use case the decomposition can be easily extended to arbitrary tensor outputs.   

\section{Conclusion}

In this work, we have presented a novel explainability technique that leverages the locally linear structure of complex neural networks to decompose the action of deep learning models into interpretable components. We have shown that the network does indeed focus on salient features if the corresponding singular vectors are combined with the input in the correct way. Furthermore, the technique can be used to dissect the network in a consistent way so that we may inspect the representations at each layer and elucidate how the operator evolves across the depth of the model. A notable example of this can be seen in the output of the convolutional layers in the vision models, which decompose the input image into spatially separated salient segments. In our view, this addresses a shortcoming in many existing explainability methods that tend to produce explanations that are a mix of all salient features. The methods that do consider intermediate representations tend to rely on interpolation and/or analysis of a subset of units. Our approach, however, outputs fine-grained explanations that consider all activated units. Next, we have shown how our method produces a natural additive symbolic representation that is equivalent across all layers, and we have outlined a method to reduce the number of contributing terms through cancellations of opposing coefficients. Finally, the main advantage of our method, we believe lies in the fact that it does not rely on axiomatic or heuristic motivations but is a principled application of established linear algebra techniques to give an accurate, interpretable characterization of complex deep learning models.     

\section{Acknowledgements}

 This work was done in the scope of the Innovate UK grant on "Explainable AI system to rationalize accelerated decision making on automotive component performance and manufacturability" [Project 10009522].

\clearpage
\bibliographystyle{unsrtnat}
\bibliography{CMRefs}

\end{document}